\documentclass[10pt, a4paper]{article}
\usepackage{misc/lrec2024}
\usepackage{times}
\usepackage{graphicx}
\usepackage{latexsym}
\usepackage[T1]{fontenc}
\usepackage[utf8]{inputenc}
\usepackage{microtype}
\usepackage{todonotes}
\usepackage{amsmath}
\usepackage{amssymb}
\usepackage{cleveref}
\usepackage{xcolor}
\usepackage{booktabs}
\usepackage{lipsum}
\usepackage{enumitem}
\usepackage{xspace}
\usepackage{listings}
\usepackage{algorithm, algorithmicx, algpseudocode}
\usepackage{soul} %
\usepackage{caption}
\usepackage{transparent} %
\usepackage{tcolorbox} %
\usepackage{multirow}
\usepackage{ulem}
\usepackage{mdframed} %
\usepackage{float}
\usepackage{fontawesome} %
\usepackage{array}
\usepackage{xfrac}
\usepackage{tipa}
\usepackage{tikz}
\usetikzlibrary{arrows.meta}

\usepackage{arevmath}

\usetikzlibrary{positioning, arrows.meta, shapes}

\definecolor{mblue}{HTML}{003F5C}
\definecolor{mviolet}{HTML}{58508D}
\definecolor{mpurple}{HTML}{BC5090}
\definecolor{msalmon}{HTML}{878787}
\definecolor{mgold}{HTML}{FFA600}
\definecolor{mdimgray}{HTML}{999999}

\crefname{lstlisting}{listing}{listings}
\Crefname{lstlisting}{Listing}{Listings}
\crefname{equ}{equation}{equations}
\Crefname{equ}{Equation}{Equations}
\Crefname{algorithm}{Algorithm}{Algorithms}

\DeclareCaptionType{equ}[Equation][List of equations]

\usepackage{courier}
\usepackage{tipa}

\usepackage{marginnote}

\definecolor{TodoColor}{rgb}{1,0.7,0.6}

\sethlcolor{TodoColor}

\newmdenv[
  linecolor=black,
  linewidth=1.2pt,
  topline=false,
  bottomline=false,
  rightline=false,
  innertopmargin=0mm,
  innerbottommargin=-0.5mm,
  innerleftmargin=1mm,
  skipabove=1.1\topsep,
  skipbelow=0.5\topsep,
]{quotebox}

\definecolor{DocumentLinkColor}{rgb}{0.4,0.6,0.3}
\newcommand{\hrefEmail}[2]{\href{mailto:#1}{\color{black}{#2}}}

\let\svthefootnote\thefootnote
\newcommand\blankfootnote[1]{%
  \let\thefootnote\relax\footnotetext{#1}%
  \let\thefootnote\svthefootnote%
}

\definecolor{ethblue}{rgb}{0,0.1,0.4}
\newcommand{\ethletter}{\hspace{-0.5mm}\text{
    \fontfamily{phv}\fontseries{bx}\fontsize{7}{\baselineskip}\selectfont
    \textit{\textbf{\color{ethblue}{E}}}}
}
\definecolor{byublue}{rgb}{0,0.1,0.2}
\newcommand{\byuletter}{\hspace{-0.5mm}\text{
    \fontfamily{phv}\fontseries{bx}\fontsize{7}{\baselineskip}\selectfont
    {\textbf{\color{byublue}{Y}}}}
}
\definecolor{cmured}{rgb}{0.5,0.1,0.1}
\newcommand{\cmuletter}{\hspace{-0.5mm}\text{
    \fontfamily{phv}\fontseries{bx}\fontsize{7}{\baselineskip}\selectfont
    {\textbf{\color{cmured}{C}}}}
}

\newcommand{\wdata}{\mathcal{W}}
\DeclareMathOperator*{\softmax}{softmax}

\title{\textsc{PWESuite}: Phonetic Word Embeddings and Tasks They Facilitate}

\name{
    Vilém Zouhar$_{\boldsymbol{=}}^{\ethletter}$ \hspace{9mm}
    Kalvin Chang$_{\boldsymbol{=}}^{\cmuletter}$ \hspace{6mm}
    Chenxuan Cui$^{\cmuletter}$ \hspace{5mm}
    Nathaniel Carlson$^{\byuletter}$ \\[0.3em]
    \bf \fontsize{11pt}{11pt}\selectfont
    Nathaniel R. Robinson$^{\cmuletter}$ \hspace{3mm}
    Mrinmaya Sachan$^{\ethletter}$ \hspace{2mm}
    David Mortensen$^{\cmuletter}$ \\[-0.2em]
}

\address{%
    $^{\ethletter}$Department of Computer Science, ETH Zurich \\
    $^{\cmuletter}$Language Technologies Institute, Carnegie Mellon University \\
    $^{\byuletter}$Department of Computer Science, Brigham Young University \\
  \texttt{
    \{\hrefEmail{vzouhar@ethz.ch}{vzouhar},\hrefEmail{msachan@ethz.ch}{msachan}\}@ethz.ch \quad
    \hrefEmail{natbcar@gmail.com}{natbcar@gmail.com} 
   } \\
  \texttt{
    \{\hrefEmail{kalvinc@cs.cmu.edu}{kalvinc},\hrefEmail{cxcui@cs.cmu.edu}{cxcui},\hrefEmail{nrrobins@cs.cmu.edu}{nrrobins},\hrefEmail{dmortens@cs.cmu.edu}{dmortens}\}@cs.cmu.edu
  } \\[-0.2em]
}

\abstract{
\vspace{2mm}
Mapping words into a fixed-dimensional vector space is the backbone of modern NLP.
While most word embedding methods successfully encode semantic information, they overlook phonetic information that is crucial for many tasks. 
We develop three methods that use articulatory features to build phonetically informed word embeddings.
To address the inconsistent evaluation of existing phonetic word embedding methods, we also contribute a task suite to fairly evaluate past, current, and future methods. 
We evaluate both (1) intrinsic aspects of phonetic word embeddings, such as word retrieval and correlation with sound similarity, and (2) extrinsic performance on tasks such as rhyme and cognate detection and sound analogies.
We hope our task suite will promote reproducibility and inspire future phonetic embedding research.
\newcommand\microspace{\hspace{0.8mm}}
\\ \newline 
\vspace{4mm}
}

\begin{document}

\maketitleabstract

\hspace{2mm}
\begin{minipage}[c]{5mm}
\includegraphics[width=\linewidth]{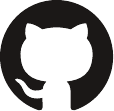}
\end{minipage}
\hspace{0.1mm}
\begin{minipage}[c]{\textwidth}
\fontsize{0.76em}{0.76em}\selectfont
Code: 
\href
{https://github.com/zouharvi/pwesuite}
{\texttt{github.com/zouharvi/pwesuite}}
\end{minipage}

\hspace{2mm}
\begin{minipage}[c]{5mm}
\includegraphics[width=\linewidth]{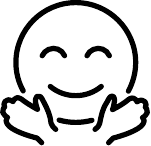}
\end{minipage}
\hspace{0.1mm}
\begin{minipage}[c]{\textwidth}
\fontsize{0.76em}{0.76em}\selectfont
Dataset: 
\begin{minipage}[c]{\textwidth}
    \href
    {https://huggingface.co/datasets/zouharvi/pwesuite-eval}
    {\texttt{huggingface.co/datasets/}} \\
    \href
    {https://huggingface.co/datasets/zouharvi/pwesuite-eval}
    {\texttt{zouharvi/pwesuite-eval}}
\end{minipage}
\end{minipage}

\blankfootnote{\hspace{-2mm}$^{=}$Co-first authors.}

\section{Introduction}

Word embeddings are omnipresent in modern NLP \citep[inter alia]{le2014distributed,pennington2014glove,almeida2019word}.
Their main benefit lies in compressing some information into fixed-dimensional vectors.
These vectors can be used as machine-learning features for NLP applications, and their study can reveal linguistic insights \citep{hamilton2016diachronic,ryskina2020where,francis2021quantifying}.
Word embeddings are often trained via methods from distributional semantics \citep{camacho2018word} and thus bear semantic information.
For example, the embedding for the word \emph{carrot} may encode higher similarity to embeddings for other vegetables than to that of \emph{ocean}.

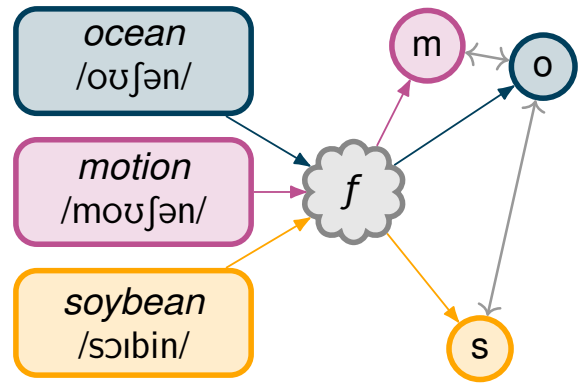
\begin{figure}[htbp]
\centering

\resizebox{\linewidth}{!}{%

  \def\boxsep{2mm}
  
  \begin{tikzpicture}[-stealth]
    \tikzset{
      word/.style = {
        rectangle,
        rounded corners = 2mm,
        text width = 2cm,
        minimum height = 1cm,
        align = center,
        line width = 1.5pt,
      },
      mot/.style = {
        draw = mpurple,
        fill = mpurple!20,
      },
      oce/.style = {
        draw = mblue,
        fill = mblue!20,
      },
      soy/.style = {
        draw = mgold,
        fill = mgold!20,
      },
      emb/.style = {
        circle,
        line width = 1.5pt,
      },
      embfunc/.style = {
        cloud,
        draw = msalmon,
        fill = msalmon!20,
        line width = 1.5pt,
      },
      blackline/.style = {
        -{Latex[round]},
        line width=0.5pt,
      }
    }
    \node[embfunc] (f) at (0,0) {$f$};
    \node[mot, word, left = 5mm of f] (motion) {\textit{motion} \\ /\textipa{moUS@n}/};
    \node[oce, word, above = \boxsep of motion] (ocean) {\textit{ocean} \\ /\textipa{oUS@n}/};
    \node[soy, word, below = \boxsep of motion] (soybean) {\textit{soybean} \\ /\textipa{sOIbin}/};
    \node[mot, emb] (mot emb) at (7mm, 14mm) {m};
    \node[oce, emb] (oce emb) at (18mm, 12mm) {o};
    \node[soy, emb] (soy emb) at (12mm, -15mm) {s};

    \draw[blackline, oce] (ocean) to (f);
    \draw[blackline, mot] (motion) to (f);
    \draw[blackline, soy] (soybean) to (f);
    \draw[blackline, oce] (f) to (oce emb);
    \draw[blackline, mot] (f) to (mot emb);
    \draw[blackline, soy] (f) to (soy emb);

    \draw[<->, mdimgray,line width=0.7pt] (oce emb) to (mot emb);
    \draw[<->, mdimgray,line width=0.7pt] (soy emb) to (oce emb);
  \end{tikzpicture}
}%
\caption{Embedding function $f$ projects words in various forms (left) to a vector space (right) such that words with a similar pronunciation (e.g., \textit{ocean} and \textit{motion}) are closer than words with a dissimilar pronunciation (e.g., \textit{ocean} and \textit{soybean}).}
\label{fig:generic_projection}
\end{figure}

Some applications may require a different type of information to be encoded.
The orthography, especially in English, can obscure the pronunciation.
A poem generation model, for instance, may need embeddings to reflect that \emph{ocean} rhymes with \emph{motion} and not with a \emph{soybean}, even though the spelling of the words' final syllables suggest otherwise (see \Cref{fig:generic_projection}). 
Such embeddings, called \textit{phonetic word embeddings}, contain phonetic information and have been of recent interest \citep{parrish2017poetic,yang2019linguistically,hu2020multilingual,sharma2021phonetic}.\footnote{The technically correct term is \textit{phonological word embeddings} but prior literature uses the term \textit{phonetic}.}
The objective is that words with similar pronunciation should be mapped to vectors near each other in embedding space.
Many tasks have benefited from incorporating phonetic word embeddings, including cognate and loanword detection \citep{rama2016siamese,nath-etal-2022-generalized,nath-etal-2022-phonetic}, named entity recognition \citep{bharadwaj2016phonologically, chaudhary2018adapting}, spelling correction \citep{zhang2021correcting}, and speech recognition \citep{phoneme2vec}. 
See \Cref{sec:what_pwe_for} for a more detailed list of possible applications.

We introduce four phonetic word embedding methods---count-based, autoencoder, and metric and contrastive learning.
Though some of these techniques are inspired by previous work, we are the first to apply them with supervision from articulatory feature vectors, a seldom-exploited form of linguistic knowledge for representation learning.

More importantly, we introduce an evaluation suite for testing the performance of phonetic embeddings.
The motivations for this are two-fold.
First, prior work is inconsistent in evaluating models.
This prevents the field from observing long-term improvements in such embeddings and from making fair comparisons across different approaches.
Secondly, when a practitioner is deciding which phonetic word embedding method to use, the go-to approach is to first apply the embeddings (generally fast) and then train a downstream model on those embeddings (compute and time intensive).
Instead, intrinsic embedding evaluation metrics (cheap)---if shown to correlate well with extrinsic metrics---could provide useful signals in embedding method selection prior to training of downstream models (expensive).
In contrast to semantic word embeddings \citep{bakarov2018survey}, we show that intrinsic and extrinsic metrics for phonetic word embeddings generally correlate with each other.
While \citet{ghannay2016evaluation} evaluate acoustic word embeddings, we specialize in phonetic word embeddings \textit{for text, not speech}.

Our main contribution is this evaluation suite for phonetic word embeddings, the equivalent of which does not yet exist in this subfield.
We also contribute multiple methods for and a survey of existing phonetic word embeddings.

\section{Survey of Phonetic Embeddings}
\label{sec:survey_phon_embeddings}
Given an alphabet $\Sigma$ and a dataset of words $\mathcal{W} \subseteq \Sigma^*$, $d$-dimensional word embeddings are given by a function $f: \mathcal{W} \rightarrow \mathbb{R}^d$. %
This function takes an element from $\Sigma^*$ (set of all possible words over the alphabet $\Sigma$) and maps it to a $d$-dimensional vector of numbers.
For many embedding functions, $\mathcal{W}$ is a finite set of words, and the embeddings are not defined for unseen words \citep{mikolov2013efficient,pennington2014glove}.
Other embedding functions---which we dub \emph{open}---are able to provide an embedding for any word $x \in \Sigma^*$  \citep{bojanowski2017enriching}.
An illustration of a \emph{phonetic} embedding function is shown in \Cref{fig:generic_projection} (\textit{motion} is closer to \textit{ocean} than to \textit{soybean}).

We use 3 distinct alphabets: characters $\Sigma_C$, \href{https://en.wikipedia.org/wiki/International_Phonetic_Alphabet}{IPA} symbols $\Sigma_P$ and \href{https://en.wikipedia.org/wiki/ARPABET}{ARPAbet} symbols $\Sigma_A$.
We use $\Sigma$ when the choice is not important and refer to elements of $\Sigma$ as characters or phonemes.
We review some \textit{semantic} embeddings in \Cref{sec:evaluation_empirical} and now focus on prior work in \emph{phonetic} embeddings.
From our formalism it also follows that we are interested in phonetic representations of \textit{textual} input.

\subsection{Poetic Sound Similarity}
\citet{parrish2017poetic} learns word embeddings capturing pronunciation similarity for poetry generation for words in the CMU Pronouncing Dictionary \citeplanguageresource{cmu-dict}. 
First, each phoneme is mapped to a set of phonetic features $\mathcal{F}$ using the function $\textsc{P2F} : \Sigma_A \rightarrow 2^\mathcal{F}$.
From the sequence of sets that each sequence of phonemes maps to, bi-grams of phonetic features are created (using Cartesian product $\times$ between sets $a_i$ and $a_{i+1}$) and counted.
The function $\textsc{CountVec}$ outputs these bi-gram counts in a vector of constant dimension. %
The resulting vector is then reduced using PCA to the target embedding dimension $d$.
\begin{align}
\hspace{-2.2mm}\textsc{W2F}(x) &= \langle \textsc{P2F}(x_i) | x_i \in x \rangle \qquad\quad \text{\textcolor{gray}{(array)}}\\
\hspace{-2.2mm}\textsc{F2V}(a) &= \textsc{CountVec.}\big(\hspace{-2mm}\bigcup_{1 \leq i \leq |a|-1} \hspace{-3mm}a_i \times a_{i+1} \big) \\
\hspace{-3mm}f_\textsc{PAR} &= \textsc{PCA}_d(\{ \textsc{F2V}(\textsc{W2F}(x)) | x \in \wdata \})
\end{align}

The function $f_\textsc{PAR}$ can provide embeddings even for words unseen during training.
This is because the only component dependent on the training data is the PCA over the vector of bigram counts, which can also be applied to new vectors.

\subsection{phoneme2vec}
\citet{phoneme2vec} do not use hand-crafted features and learn phoneme embeddings using a more complex, deep-learning, model.
They start with a gold sequence of phonemes $(x_i)$ and a noisy sequence of phonemes $(y_i)$.
The phonemes are one-hot encoded in matrices $X$ and $Y$.
The gold sequence is first read by an LSTM model, yielding the initial hidden state $h_0$.
From this hidden state, the phonemes $(\hat{y_i})$ are decoded using teacher forcing (upon predicting $\hat{y_i}$, the model receives the correct $x_i$ as the input).
The phoneme embedding matrix $V$ is trained jointly with the model weights and constitutes the embedding function. %
\begin{flalign}
&\hspace{-1mm}h_0 = \textsc{LSTM}(XV) \\
&\hspace{-1mm}\mathcal{L}_\text{p2v} = 
    -\hspace{-4mm}\sum_{0 < i \leq |y|} \hspace{-2mm}\log  \softmax( \textsc{LSTM}(Y_{<i} V)_{y_i} )\hspace{-1.8mm}
\end{flalign}

For a fair comparison, we average these vectors which are \textit{phoneme}-level to get word-level embeddings.
In addition, in contrast to other embeddings, these phoneme embeddings are only 50-dimensional.
We revisit the question of dimensionality in \Cref{sec:dimensionality}.

\subsection{Phonetic Similarity Embeddings}

\citet{sharma2021phonetic} propose a vowel-weighted phonetic similarity metric to compute similarities between words.
They then use it for training phonetic word embeddings which should share some properties with this similarity function.
This is in contrast to the previous approaches, where the embedding training is indirect, on an auxiliary task.
Given a sound similarity function $S_\text{PSE}$, they construct a matrix of similarity scores $S\in \mathbb{R}^{|\wdata|\times |\wdata|}$ such that $S_{i,j} = S_\text{PSE}(\mathcal{W}_i, \mathcal{W}_j)$.
On this matrix, they use non-negative matrix factorization to learn the embedding matrix $V\in \mathbb{R}^{|\wdata|\times d}$ such that the following loss is minimized:
\begin{equation}
    \mathcal{L}_\text{PSE} = ||S - V\cdot V^T ||^2
\end{equation}

Then, the $i$-th row of $V$ contains the embedding for $i$-th word from $\wdata$.
A critical disadvantage of this approach is that it cannot be used for embedding new words because the matrix $V$ would need to be recomputed again.
We apply the sound similarity function $S_\text{PSE}$, defined specifically for English, to all evaluation languages.

\section{Our Models}
\label{sec:our_model}
We now introduce several embedding baselines.
Then, we describe our articulatory distance metric and models trained with supervision therefrom.

\subsection{Count-based Vectors}

Perhaps the most straightforward way of creating a vector representation for a sequence of input characters or phonemes $x \in \Sigma^*$ is simply counting n-grams in this sequence.
We use a term frequency-inverse document frequency (TF-IDF) vectorizer of 1-, 2-, and 3-grams (formally denoted $[x]_n$) across the input sequence of symbols (e.g. characters) with a maximum of 300 features.
This vector then becomes our word embedding.
For instance, the first dimension may be the TF-IDF score or occurrence count of the bigram $\langle$/\textipa{dIn}/, /\textipa{a}/$\rangle$.
\begin{align}
\hspace{-2mm}\textsc{C2V}(x) &= [x]_1 \cup [x]_2 \cup [x]_3 \qquad \textcolor{gray}{\text{(features)}}\\
\hspace{-2mm}f_\text{count}(x) &= \textsc{TF-IDF}_{{\substack{\text{feat}\hspace{0.5mm}\\\text{ures}}}=d}(\{\textsc{C2V}(x) | x \in \wdata \})\hspace{-2mm}
\end{align}
\vspace*{-3mm}

\subsection{Autoencoder}
Another common approach, though less interpretable, for vector representation with fixed dimension size is an encoder-decoder autoencoder.
Specifically, we use this architecture together with forced-teacher decoding and use the bottleneck vector as the phonetic word embedding.
In an ideal case, the fixed-size bottleneck contains all the information to reconstruct the whole sequence from $\Sigma^*$.
\begin{align}
\hspace{-3mm}
f_\theta(x) = \textsc{LSTM}(x|\theta) \,\qquad\qquad \textcolor{gray}{\text{(encoder)}}\\
\hspace{-3mm}
d_{\theta'}(x) = \textsc{LSTM}(x|\theta') \qquad\qquad \textcolor{gray}{\text{(decoder)}} \\
\hspace{-2.5mm}
\mathcal{L}_\text{auto.} \hspace{-0.5mm}=\hspace{-3mm}\sum_{0 < i \leq |x|}\hspace{-3mm} -\log \text{softmax}(d_{\theta'}(f_\theta(x)|x_{<i})_{x_i} )
\end{align}

\subsection{Phonetic Word Embeddings With Articulatory Features}
\label{sec:learning_phonetic_embeddings}

\subsubsection{Articulatory Features and Distance}
\label{sec:articulatory_distance}
Articulatory features \citep{bloomfield1933,jakobson1951preliminaries,chomsky1968sound} decompose sounds into their constituent properties.
Each segment can be mapped to a vector with $n$ different features (24 for PanPhon \citealp{mortensen-etal-2016-panphon}) such as whether the phoneme segment is produced with a nasal airflow or if it is produced with raised or lowered tongue tip.
A segment is a group of phonetic characters (e.g., as defined by Unicode) that represent a single sound.
We define $a{:}\, \Sigma_P \rightarrow \{-1, 0, +1\}^{24}$ as the function which maps a phoneme segment into a vector of articulatory features.
Values +1/-1 mean present/not present and the value 0 is used when the feature is irrelevant.

The articulatory distance, also called \textit{feature edit distance} \citep{mortensen-etal-2016-panphon}, is a version of Levenshtein distance with custom costs.
Specifically, the substitution cost is proportional to the Hamming distance between the source and target when they are represented as articulatory feature vectors.
Omitting edge-cases, it is defined as:

\hspace{-3mm}\begin{minipage}{\linewidth}
\begin{align}
\text{}
\end{align}
\vspace{-17mm}
\begin{align}
    A_{i,j}(x, x') &= \min \left\{\hspace{-2mm}
    \begin{array}{l}
      A_{i - 1, j}(x, x') + d(x)\\
      A_{i, j - 1}(x, x') + i(x')\\
      A_{i - 1, j - 1}(x, x') + s(x_i, x'_j)\\
    \end{array}\right.\nonumber \\
 A(x,x') &= A_{|x|,|x'|}(x, x')
\end{align}
\end{minipage}
\vspace{4mm}

where $d$ and $i$ are deletion and insertion costs, which we set to constant $1$.
The function $s$ is a substitution cost, defined as the number of elements (normalized) that need to be changed to render the two articulatory vectors identical:
\begin{equation}
  \label{eq:2}
  s(x, x') = \frac{1}{24}\sum_{i=1}^{24} |a(x)_i - a(x')_i|
\end{equation}

The articulatory distance $A$ induces a metric space-like structure for words in $\Sigma^*$.
It quantifies the phonetic similarity between a pair of words, capturing the intuition that /pæt/ and /bæt/ are phonetically closer than /pæt/ and /hæt/, for example.

\subsubsection{Metric Learning}
As one means of generating word embeddings, we use the last hidden state of an LSTM-based model.
We use characters $\Sigma_C$, IPA symbols $\Sigma_P$ (\Cref{sec:survey_phon_embeddings}) and articulatory feature vectors as the input.
We discuss these choices and especially their effect on performance and transferability in \Cref{sec:language_transfer}.

We now have a function $f$ that produces a vector for each input word.
However, it is not yet trained to produce vectors encoding phonetic information.
We, therefore, define the following differentiable loss where $A$ is the articulatory distance.

\text{}\begin{align}
&\mathcal{L}_\text{dist.} = \frac{1}{|\mathcal{W}|}\sum_{\substack{x_a\in\mathcal{W}\\x_b\sim \mathcal{W}}} \Big(\,||f_\theta(x_a)- f_\theta(x_b)||^2 \hspace{13mm}\nonumber 
\end{align}
\text{}\vspace{-13mm}
\begin{align}
&\hspace{46mm}  - A(x_a,x_b) \Big)^2
\hspace{-2mm}
\end{align}
This forces the embeddings to be spaced in the same way as the articulatory distance ($A$, \Cref{sec:articulatory_distance}) would space them.
Metric learning (learning a function to space output vectors similarly to some other metric) has been employed previously \citep{yang2006distance,bellet2015metric,kaya2019deep} and was used to train \textit{acoustic} embeddings by \citet{yang2019linguistically}.

\subsubsection{Triplet Margin loss}

While the previous approach forces the embeddings to be spaced exactly as by the articulatory distance function $A$, we may relax the constraint so only the structure (ordering) is preserved.
This is realized by triplet margin loss:
\begin{equation}\label{trip:1}
    \mathcal{L}_\text{triplet} = \max \begin{cases}
        0 \\
        \alpha + |f_\theta(x_a)-f_\theta(x_p)| \\
        \quad - |f_\theta(x_a)-f_\theta(x_n)|
    \end{cases}
\end{equation}

We consider all possible ordered triplets of distinct words $(x_a, x_p, x_n)$ such that $A(x_a,x_p) < A(x_a,x_n)$.
We refer to $x_a$ as the anchor, $x_p$ as the positive example, and $x_n$ as the negative example.
We then minimize $\mathcal{L}_\text{triplet}$ over all valid triplets.
This allows us to learn $\theta$ for an embedding function $f_\theta$ that preserves the local neighbourhoods of words defined by $A(x,x^{\prime})$.
In addition, we modify the function $f_\theta$ by applying attention to all hidden states extracted from the last layer of the LSTM encoder.
This allows our model to focus on phonemes that are potentially more useful when trying to summarize the phonetic information in a word. A related approach was used by \citet{yang2019linguistically} to learn acoustic word embeddings.
Although contrastive learning is a more intuitive approach, it yielded only negative results: \scalebox{0.89}{$\left(\text{exp}(|f_\theta(x_a)-f_\theta(x_p)|^2)\right)$/$\left(\sum \text{exp}(|f_\theta(x_a)-f_\theta(x_n)|^2)\right)$}.

Though metric learning and triplet margin loss have been applied previously to similar applications, we are the first to apply them using articulatory features and articulatory distance.

\subsection{Phonetic Language Modeling}

To shed more light into the true landscape of phonetic word embedding models, we describe here a model which did not perform well on our suite of tasks (in contrast to other models).
A common way of learning word embeddings now is to train on the masked language model objective, popularized by BERT \citep{devlin2019bert}. We input articulatory features from PanPhon into several successive Transformer \citep{transformer} encoder layers and a final linear layer that predicts the masked phone.
Positional encoding is added to each input. 
We prepend and append \texttt{[CLS]} and \texttt{[SEP]} tokens, respectively, to the phonetic transcriptions of each word, before we look up each phone's PanPhon features.
Unlike BERT, we do not train on the next sentence prediction objective. %
As such, we use mean pooling to extract a word-level representation instead of [CLS] pooling. 
In addition, we do not add an embedding layer because we are not interested in learning individual phone embeddings but rather wish to learn a word-level embedding.
Unlike metric learning and the triplet margin loss, there is no explicit objective to incorporate the pronunciation similarity, which may explain the underperformance of this model.

\section{Evaluation Suite (key contribution)}
\label{sec:evaluation_suite}

We now introduce the embedding evaluation metrics of our suite, the primary contribution of this paper.
We draw inspiration from evaluating semantic word embeddings \citep{bakarov2018survey} and work on phonetic word embeddings \citep{parrish2017poetic}.
In some cases, the distinction between intrinsic and extrinsic evaluations is tenuous (e.g., retrieval and analogies).
The main characteristic of intrinsic evaluation is that they are efficiently computed and are not part of any specific application.
In contrast, extrinsic evaluation metrics directly measure the usefulness of the embeddings for a particular task.

We evaluate with 9 phonologically diverse languages: Amharic,\hspace{-1mm}$^*$ Bengali,\hspace{-1mm}$^*$ English, French, German, Polish, Spanish, Swahili, and Uzbek.
Languages marked with $*$ use non-Latin script.
The non-English data (200k tokens each) is from CC-100 \citeplanguageresource{wenzek2020ccnet,conneau2020unsupervised}, while the English data (125k tokens) is from the CMU Pronouncing Dictionary \citeplanguageresource{cmu-dict}. 

\subsection{Intrinsic Evaluation}

\subsubsection{Articulatory Distance}

The unifying desideratum for phonetic embeddings is that they should capture the concept of pronunciation similarity.
Recall from \Cref{sec:survey_phon_embeddings} that phonetic word embeddings are a function $f: \Sigma^* \rightarrow \mathbb{R}^d$.
In the vector space of $\mathbb{R}^d$, there are two widely used notions of similarity $S$.
The first is the \emph{negative $L_2$ distance} and the other is the \emph{cosine similarity}.
Consider three words $x, x'$ and $x''$.
Using either metric, $S(f(x), f(x'))$ yields the embedding similarity between $x$ and $x'$.
On the other hand, since we have prior notions of similarity $S_P$ between the words, e.g., based on a rule-based function, we can use this to represent the similarity between the words: $S_P(x, x')$.
We want to have embeddings $f$ such that $S {\circ} f$ produces results close to $S_P$.
There are at least two ways to verify that the similarity results are close.
First is exact equality.
For example, if $S_P(x, x') = 0.5, S_P(x, x'') = 0.1$, we want $S(f(x), f(x')) = 0.5, S(f(x), f(x'')) = 0.1$.
We can measure this using Pearson's correlation coefficient between $S\circ f$ and $S_P$.
On the other hand, we may consider only the relative similarity values.
Following the previous example, we would only care that $S(f(x), f(x')) > S(f(x), f(x''))$.
In this case we use Spearman's correlation coefficient between $S\circ f$ and $S_P$.
For the rule-based similarity metric $S_P$, we use \emph{articulatory distance} \citep{mortensen-etal-2016-panphon}, as described in \Cref{sec:articulatory_distance}.
For computation reasons, we randomly sample 1000 pairs.

\begin{table*}[htbp]
\renewcommand\arraystretch{1.24}
\centering
\resizebox{\linewidth}{!}{
\begin{tabular}{cl>{\hspace{-5mm}}ccc<{\hspace{4mm}}ccc<{\hspace{2mm}}c}
\toprule
&& \multicolumn{3}{c}{\textbf{\textsc{Intrinsic}}\hspace{5mm}\text{}} & \multicolumn{3}{c}{\textbf{\textsc{Extrinsic}}} & \textbf{\textsc{Overall}} \\ 
& \textbf{Model} & \textbf{Human Sim.} & \textbf{Art. Dist.} & \textbf{Retrieval} & \textbf{Analogies} & \textbf{Rhyme} & \textbf{Cognate} & \\
&& (Pearson) & (Pearson) & (rank perc.) & (Acc@1) & (accuracy) & (accuracy) & \\
\midrule
\parbox[t]{2mm}{\multirow{4}{*}{\rotatebox[origin=c]{90}{Ours}}}
& Metric Learner 
    & 0.46 & 0.94 & 0.98
    & 84\% & 83\% & 64\%
    & 0.78  \\
& Triplet Margin
    & 0.65 & 0.96 & 1.00
    & 100\% & 77\% & 66\%
    & 0.84 $\star$\hspace{-3mm}\text{}\\
& Count-based 
    & 0.82 & 0.10 & 0.84
    & 13\% & 79\% & 68\%
    & 0.56 \\
& Autoencoder
    & 0.49 & 0.16 & 0.73
    & 50\% & 61\% & 50\%
    & 0.50 \\[2pt]
\cmidrule{2-2}
\parbox[t]{2mm}{\multirow{3}{*}{\rotatebox[origin=c]{90}{Others'}}}
& Poetic Sound Sim.
    & 0.74 & 0.12 & 0.78
    & 35\% & 60\% & 57\% 
    & 0.53 \\
& phoneme2vec 
    & 0.77 & 0.09 & 0.80
    & 17\% & 88\% & 64\%
    & 0.56 \\
& Phon. Sim. Embd. 
    & 0.16 & 0.05 & 0.50
    & 0\% & 51\% & 52\%
    & 0.29 \\[2pt]
\cmidrule{2-2}
\parbox[t]{2mm}{\multirow{4}{*}{\rotatebox[origin=c]{90}{Semantic}}}
& BPEmb
    & 0.23 & 0.08 & 0.60
    & 5\% & 54\% & 66\%
    & 0.36 \\
& fastText
    & 0.25 & 0.12 & 0.64
    & 2\% & 58\% & 68\%
    & 0.38 \\
& BERT
    & 0.10 & 0.34 & 0.69
    & 4\% & 58\% & 63\% 
    & 0.40 \\ 
& INSTRUCTOR
    & 0.60 & 0.12 & 0.73
    & 7\% & 54\% & 66\% 
    & 0.45 \\ 
\bottomrule
\end{tabular}
}
\caption{Embedding method performance in our evaluation suite. Higher number is always better.
}
\label{tab:evaluation_all}
\end{table*}

\subsubsection{Human Judgement}

\citet{vitz1973predicting} asked people to judge the sound similarity of English words.
For selected word pairs, we denote the collected judgements (scaled from 0--least similar to 1--identical) with the function $S_H$.
For example, $S_H(\textit{slant}, \textit{plant)} = 0.9$ and $S_H(\textit{plots}, \textit{plant}) = 0.4$.
Like the previous task, we find correlations between $S {\circ} f$ and $S_H$.
We note $S_H$ judgments were produced from a small English-only corpus.
These limitations highlight the importance of including analyses with $A$, rather than $S_H$ alone.
In fact, $A$ and $S_H$ do not correlate positively, with Pearson coefficient $-0.74$.

\subsubsection{Retrieval}

An important usage of word embeddings is the retrieval of associated words, which is also  utilized in the analogies extrinsic evaluation and other applications.
Success in this task means that the new embedding space has the same local neighbourhood as the original space induced by some non-vector-based metric.
Given a word dataset $\wdata$ and one word $w \in \wdata$, we sort $\wdata \setminus \{w\}$ based on both $S {\circ} f$ and $S_P$ distance from $w$.
Based on this ordering, we define the immediate neighbour of $w$ based on $S_P$, denoted $w_N$ and ask the question \emph{What is the average rank of $w_N$ in the ordering by $S {\circ} f$?}
If the similarity given by $S {\circ} f$ is copying $S_P$ perfectly, then the rank will be 0 because $w_N$ will be the closest to $w$ in $S {\circ} f$.

Again, for $S_P$ we use the articulatory distance $A$ (\Cref{sec:articulatory_distance}).
Even though there are a variety of possible metrics to evaluate retrieval, we focus on the average rank.
We further cap the retrieval neighborhood at $n=1000$ samples and compute percentile rank as $\frac{n-r}{n}$.
This choice is done so that the metric will be bounded between 0 (worst) and 1 (best), which will become important for overall evaluation later (\Cref{sec:overall_score}).

\paragraph{Error analysis.}
\label{sec:retrieval_error_analysis}

We identify two types of errors in the retrieval task for the \textit{Metric Learner} model with articulatory features.
The first one are simply incorrect neighbours with low sound similarity, such as the word \textit{carcass}, whose correct neighbour is \textit{cardiss} but for which \textit{krutick} is chosen.
The next group are plausible ones, such as for the word \textit{counterrevolutionary}, its neighbour in articulatory distance space \textit{counterinsurgency} and the retrieved word \textit{cardiopulmonary}.
In this case we might even say that the retrieved word is closer.

\subsection{Extrinsic Evaluation}

\subsubsection{Rhyme Detection}

There are multiple types of word rhymes, most of which are based around two words sounding similarly.
We focus on perfect rhymes: when the sounds from the last stressed syllables are identical.
An example is \textit{grown} and \textit{loan}, even though the surface character form does not suggest it.
Clearly, this task can be deterministically solved if one has access to the articulatory and stress information of the concerned words.
Nevertheless, we wish to evaluate whether this information can be encoded in a fixed-length vector produced by $f$.
We create a balanced binary prediction task for rhyme detection in English and train a small multi-layer perceptron classifier on top of pairs of word embeddings.
The linking hypothesis is that the higher the accuracy, the more useful information for the task there is in the embeddings.

\subsubsection{Cognate Detection}

Cognates are words in different languages that share a common origin.
We include \textit{loanwords} alongside genetic cognates.
Similarly to rhyme detection, we frame cognate detection as a binary classification task where the input is a potential cognate pair.
CogNet \cite{batsuren-etal-2019-cognet} is a large cognate dataset of many languages, making it ideal to evaluate the usefulness of phonetic embeddings. 
We add non-cognate, distractor pairs in the dataset by finding the orthographically closest word that is not a known cognate.
For example, \textit{plant}$_\text{EN}$ and \textit{plante}$_\text{FR}$ are cognates, while \textit{plant}$_\text{EN}$ and \textit{plane}$_\text{EN}$ are not.
Although cognates also preserve some of the similarities in the meaning, we detect them using phonetic characteristics only.

\subsubsection{Sound Analogies}

Just as distributional semantic vectors can complete word-level analogies such as \textit{man : woman $\leftrightarrow$ king : queen} \citep{word2vec-analogies}, so too should well-trained phonetic word embeddings capture sound analogies.
For example of a sound analogy, consider /\textipa{dIn}/ : /\textipa{tIn}/ $\leftrightarrow$ /\textipa{zIn}/ : /\textipa{sIn}/.
The difference within the pairs is [$\pm$voice] in the first phoneme segment of each word. 

With this intuition in mind, we define a \textit{perturbation} as a pair of phonemes $(p, q)$ differing in one articulatory feature.
We then create a sound analogy corpus of 200 quadruplets $w_1 : w_2 \leftrightarrow w_3 : w_4$ for each language, with the following procedure:
\begin{enumerate}[itemsep=0.0pt]
    \item Choose a random word $w_1 \in \mathcal{W}$ and one of its phonemes on random position $i$: $p_1 = w_{1,i}$.
    \item Randomly select two perturbations of the same phonetic feature so that $p_1 : p_2 \leftrightarrow p_3 : p_4$, for example /t/ : /d/ $\leftrightarrow$ /s/ : /z/.
    \item Create $w_2$, $w_3$, and $w_4$ by duplicating $w_1$ and replacing $w_{1, i}$ with $p_2$, $p_3$, and $p_4$.
    The new words $w_2, w_3$, and $w_4$ do not have to be a real word in the language but we are still interested in analogies in the space of all possible words and their detection. This is possible only for \textit{open} embeddings.
\end{enumerate}

We apply the above procedure 1 or 2 times to create 200 analogous quadruplets with 1 or 2 perturbations (evenly split).
We then measure the Acc@1 to retrieve $w_4$ from $\mathcal{W} \cup \{ w_4 \}$.
We simply measure how many often the closest neighbour of $w_2 - w_1 + w_3$ is $w_4$.
Our analogy task is different from that of \citet{parrish2017poetic} who focused on morphological derivation\footnote{Example \textit{decide : decision $\leftrightarrow$ explode : explosion}.}
and that of \citet{silfverberg-etal-2018-sound}, which show that phoneme embeddings learned via the word2vec objective demonstrate sound analogies at the \textit{phoneme} level. 
We consider sound analogies at the \textit{word} level.

\subsection{Overall Score}
\label{sec:overall_score}

Since all the measured metrics are bounded between 0 and 1, we can define the \textit{overall} score for our evaluation suite as the arithmetic average of results from each task.
We mainly consider the results of all available languages averaged but later in \Cref{sec:language_transfer} discuss results per language as well.
To allow for future extensions in terms of languages and tasks, this evaluation suite is versioned, with the version described in this paper being \texttt{v1.0}.

\section{Evaluation}
\label{sec:evaluation_empirical}

We now compare all the aforementioned embedding models using our evaluation suite.
We show the results in \Cref{tab:evaluation_all} with three categories of models.
Our models trained using some articulatory features or distance supervision (\Cref{sec:our_model}) are given first, followed by other phonetic word embedding models (\Cref{sec:survey_phon_embeddings}).
We also include non-phonetic word embeddings, not as a fair baseline for comparison but to show that these embeddings are different from phonetic word embeddings and are not suited for our tasks: fastText \citeplanguageresource{grave2018learning}, BPEmb \citeplanguageresource{heinzerling2018bpemb}, BERT \citep{devlin2019bert} and INSTRUCTOR \citep{instructor}.
We chose these embeddings because they are open (i.e., they provide embeddings even to words unseen in the training data).
All of these embeddings except for BERT and INSTRUCTOR are 300-dimensional
(see \Cref{sec:dimensionality}).

\begin{figure}[htbp]
\centering
\includegraphics[width=\linewidth]{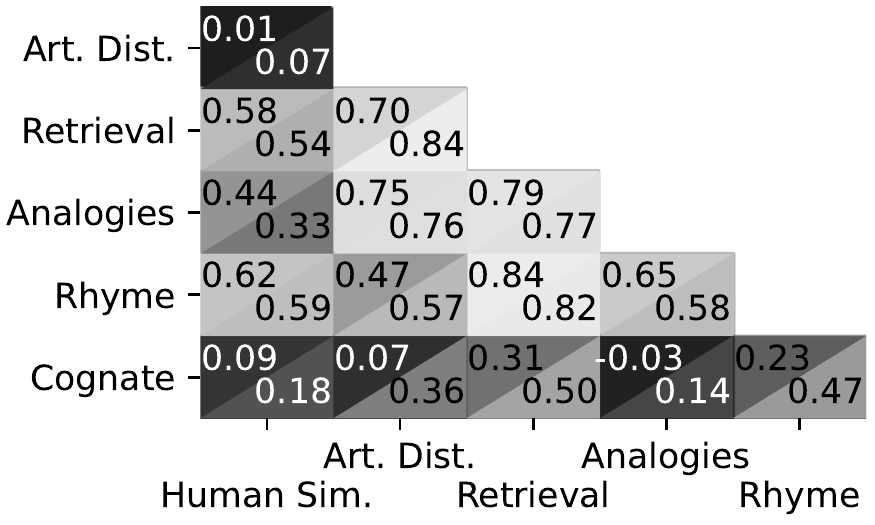}
\caption{Spearman (upper left) and Pearson (lower right) correlations between performance on suite tasks. All models from \Cref{tab:evaluation_all} are used.}
\label{fig:task_correlation}
\end{figure}

\subsection{Model Comparison}
\label{sec:model_comparison}

In \Cref{tab:evaluation_all} we show the performance of all previously described models.
The \textit{Triplet Margin} model is best overall, outperforming \textit{Metric Learner}, despite its less direct supervision in training.
However, it also requires the longest time to train.\footnote{The overall GPU budget for all included experiments is 100 hours on GTX 1080 Ti. We include reproducibility details in the code repository.}
Surprisingly, the best model for human similarity is a simple \textit{count-based} model.
Semantic word embeddings perform worse than explicit phonetic embeddings, most notably on human similarity and analogies.
However, they do perform reasonably on cognate detection.

We now examine how much the performance on one task (particularly an intrinsic one) is predictive of performance on another task.
We measure this across all systems in \Cref{tab:evaluation_all} and revisit this topic later for creating variations of the same model.
For lexical/semantic word embeddings, \citet{bakarov2018survey} notes that the individual tasks \textit{do not correlate} among each other.
In \Cref{fig:task_correlation}, we find the contrary for some of the tasks (e.g., retrieval-rhyme or retrieval-analogies).
Importantly, there is no strong negative correlation between any tasks, suggesting that performance on one task is not a tradeoff with another. 

\begin{table}[htbp]
\centering
\resizebox{0.85\linewidth}{!}{
\begin{tabular}{lccc}
\toprule
\textbf{Model} & \textbf{Art.} & \textbf{IPA} & \textbf{Text} \\
\midrule
Metric Learner & 0.78 & 0.64 & 0.62 \\
Triplet Margin & 0.84 & 0.84 & 0.79 \\
Autoencoder & 0.50 & 0.41 & 0.41 \\
Count-based & - & 0.56 & 0.51 \\
\bottomrule
\end{tabular}
}
\caption{Overall performance of models with various input features. Art. = articulatory features.}
\label{tab:input_features}
\end{table}

\subsection{Input Features}

For all of our models, it is possible to choose the input feature type, which has an impact on the performance, as shown in \Cref{tab:input_features}.
Unsurprisingly, the more phonetic the features are, the better the resulting model is.
In the \textit{Metric Learner} and \textit{Triplet Margin} models we are still using supervision from articulatory distance, and despite that, the input features play a major role.

\begin{figure}[t]
\centering
\includegraphics[width=\linewidth]{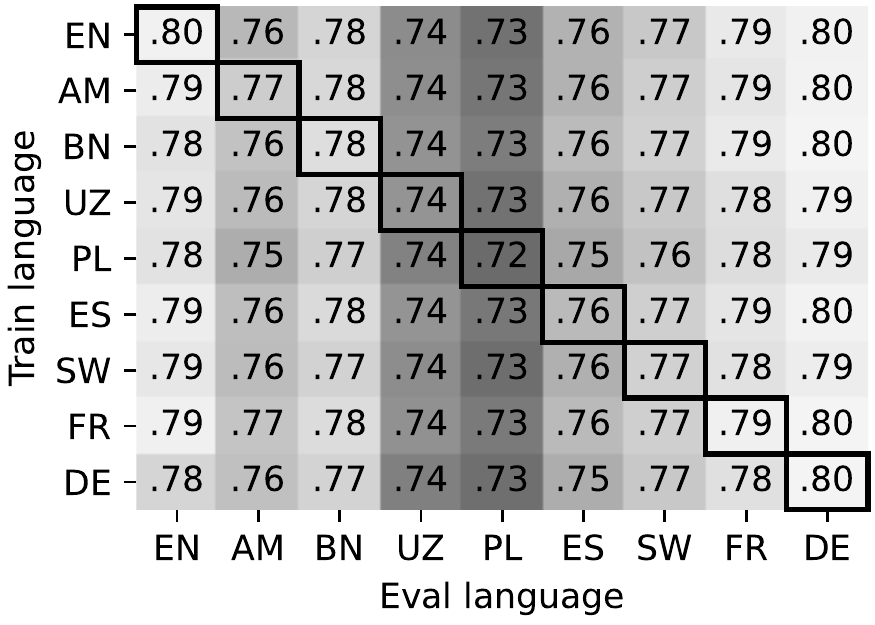}
\caption{Suite score of Metric Learner with articulatory features trained on one language and evaluated on another one. Diagonal shows models trained and evaluated on the same language.}
\label{fig:language_mismatch}
\end{figure}

\subsection{Transfer Between Languages}
\label{sec:language_transfer}

Recall from \Cref{sec:learning_phonetic_embeddings} that there are multiple feature types that can be used for our phonetic word embedding model: orthographic characters, IPA characters and articulatory feature vectors.
It is not surprising that the characters as features provide little transferability when the model is trained on a different language than it is evaluated on.
The transfer between languages for a different model type, shown in \Cref{fig:language_mismatch}, demonstrates that not all languages are equally challenging (e.g. Polish is more challenging than German).
Furthermore, the articulatory features appear to be very useful for generalizing across languages.
This echoes the findings of \citet{li-phone-embeddings}, who also break down phones into articulatory features to share information across, possibly unseen, phones.

\subsection{Embedding Topology Visualization}

\begin{figure}[htbp]
\includegraphics[width=\linewidth]{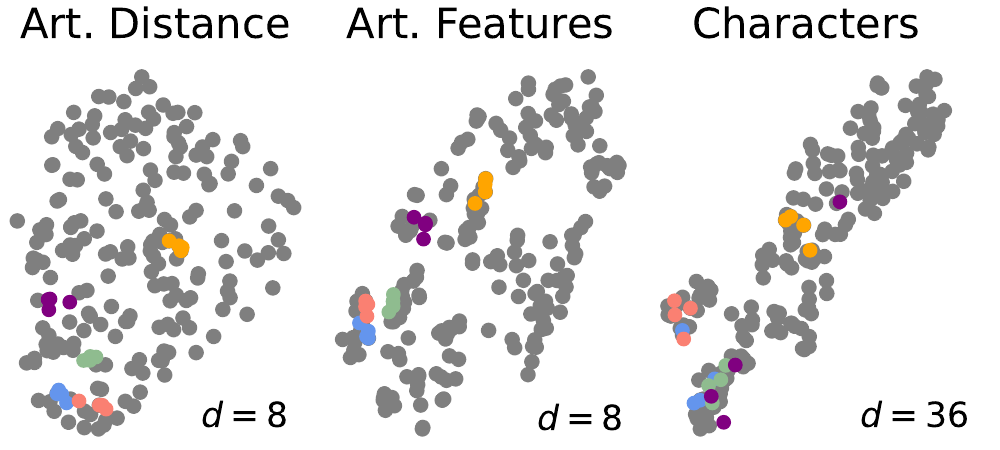}
\vspace{1mm}
\caption{T-SNE projection of articulatory distance and embedding spaces from the metric learning models with articulatory or character features. Each point corresponds to one English word. Differently coloured clusters were selected in the articulatory distance space (left) and highlighted in other spaces. $d$ is the average distance within the clusters normalized with average distance between points (unitless). Articulatory Features (center) result in tighter clusters than Characters (right).}
\label{fig:clusters}
\end{figure}

The differences between feature types in \Cref{tab:input_features} may not appear very large.
Closer inspection of the clusters in the embedding space in \Cref{fig:clusters} reveals, that using the articulatory feature vectors or IPA features yields a vector space which resembles one induced by the articulatory distance the most.
This is in line with $A$ (articulatory distance, \Cref{sec:articulatory_distance}) being calculated using articulatory features and is used for the model supervision.

\subsection{Dimensionality and Train Data Size}
\label{sec:dimensionality}

\begin{figure}[htbp]
\centering
\includegraphics[width=0.93\linewidth]{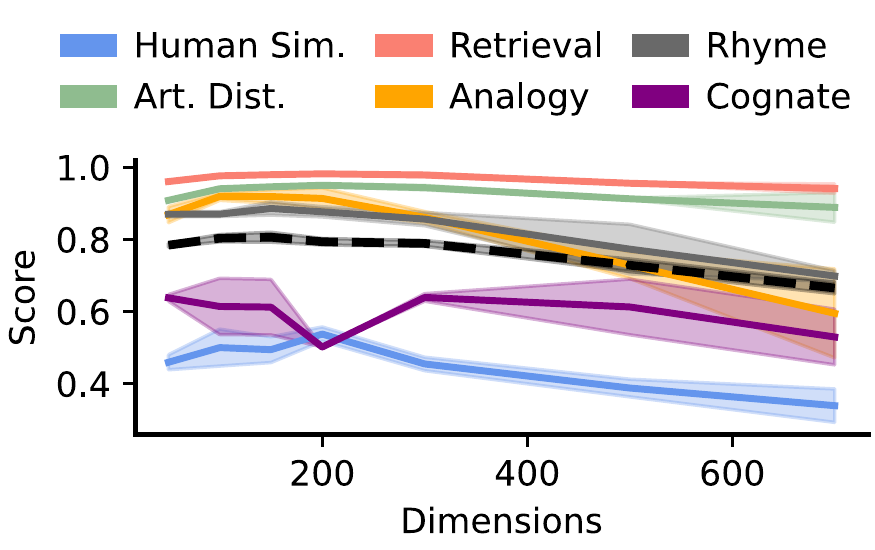}

\includegraphics[width=\linewidth,trim=0cm 0cm 0cm 1.5cm,clip]{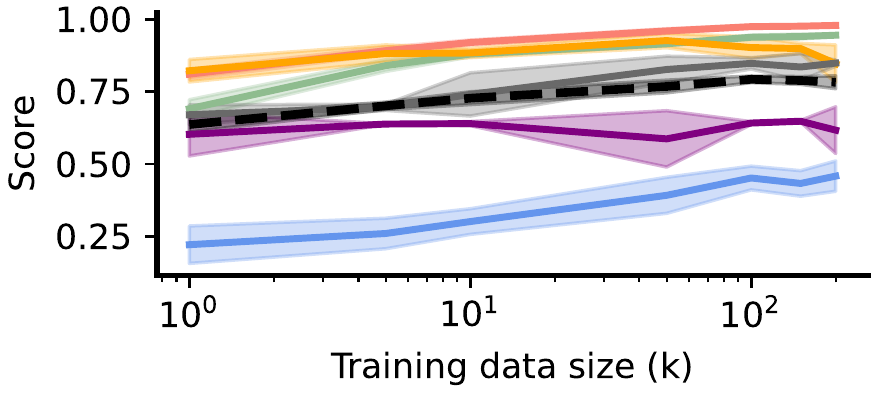}
\caption{Metric Learner performance with \textbf{varying dimensionality} (top) and \textbf{varying training data size} (bottom) with articulatory features. Bands show 95\% confidence intervals from t-distribution.}
\label{fig:dimensions_perf}
\end{figure}

So far we used 300-dimensional embeddings.
This choice was motivated solely by the comparison to other word embeddings.
Now we examine how the choice of dimensionality, keeping all other things equal, affects individual task performance.
The results in \Cref{fig:dimensions_perf} (top) show that neither too small nor too large a dimensionality is useful for the proposed tasks.
Furthermore, there is little interaction between the task type and dimensionality.
As a result, model ranking based on each task is very similar across dimensions, with Spearman and Pearson correlations of $0.61$ and $0.79$, respectively.

A natural question is how data-intensive the proposed metric learning method is.
For this, we constrained the training data size and show the results in \Cref{fig:dimensions_perf} (bottom).
Similarly to changing the dimensionality, the individual tasks react to changing the training data size without an effect of the task variable.
The Spearman and Pearson correlations are $0.64$ and $0.65$, respectively.

\section{Discussion}
\label{sec:discussion}
\subsection{The Field of Phonology}

Phonological features, especially articulatory features, have played a strong role in phonology since \citet{bloomfield1933} and the work of Prague School linguists \citep{trubetskoy1939grundzuge,jakobson1951preliminaries}.
The widely used articulatory feature set employed by PanPhon originates in the monumental \textit{Sound Pattern of English} \cite{chomsky1968sound}, which assumes a universal set of discrete phonological features and that all speech sounds in all languages consist of vectors of these features.
The similarity between these feature vectors should capture the similarity between sounds.
This position is born out in our results.
These features encode a wealth of knowledge gained through decades of linguistic research on how the sound systems of languages behave, both synchronically and diachronically.
While there is evidence that phonological features are emergent rather than universal \cite{mielke2008emergence}, these results suggest they can nevertheless contribute robustly to computational tasks.
Phonetic word embeddings also represent more closely how humans and, in particular, children, interact with language (through sound rather than abstract meaning).
Their study may have further applications in the fields of phonetics and phonology.

\subsection{Applications}
\label{sec:what_pwe_for}

Phonetic word embeddings are more ``niche'' than their semantic counterparts but there are many applications shown to benefit from them.
\begin{itemize}[left=0mm]
\item \textbf{Cognate/loanword detection} \citep{rama2016siamese,nath-etal-2022-generalized,nath-etal-2022-phonetic}.
    Along with semantic similarity, phonetic similarity measured in some latent transformation of articulatory features suggests cognacy or lexical borrowing.
\item \textbf{Multilingual named entity recognition} \citep{bharadwaj2016phonologically,chaudhary2018adapting}.
    Learning word embeddings from PanPhon features enables cross-lingual transfer for named entity recognition since named entities will likely bear pronunciation similarities across languages.
\item \textbf{Keyphrase extraction} \citep{ray2019keyphrase,alotaibi2022keyphrase}.
    Keyphrase extraction from Tweets for disaster relief can leverage PanPhon features to take advantage of the tendency for orthographic variants of the same word across different Tweets to share similar pronunciations. 
\item \textbf{Spelling correction} \citep{tan2020spelling,zhang2021correcting}.
    Imbuing word embeddings with pronunciation similarity helps in correcting typing mistakes by substituting words with their phonetic transcription and similar-sounding words.
    Another approach is to pretrain a spelling-correction model on phonetic units.
\item \textbf{Phonotactic learning} \citep{mirea2019using,romero2021use}.
    Phonetic information is a necessary part in deriving phonotactic patterns and vector representations.
\item \textbf{Multimodal word embeddings} \citep{zhu2020learning,zhu2021incorporating}.
    Phonetic and syntactic information can be incorporated into semantic word embeddings.
\item \textbf{Spoken language understanding} \citep{8639553,Chen_2021,phoneme2vec}.
    Training with phoneme embeddings can reduce errors from confusing phonetically similar words in automatic speech recognition so that such errors do not propagate to downstream natural language understanding tasks. 
\item \textbf{Language identification} \citep{electronics10182259,salesky2021sigtyp}
    Phonological features help in distinguishing between languages and their identification.
\item \textbf{Poetry generation} \citep{9364588,yi2018chinese}
    Word sounds and their pronunciations are critical for poetry and incorporation of this information helps in automatic poetry generation.
\item \textbf{Linguistic analysis} \citep{hamilton2016diachronic,ryskina2020where,francis2021quantifying}
    Apart from direct applications, there exist many investigations and analyses on what phonological and phonetic features are encoded by speakers.
    Phonological word embeddings are one tool by which this can be studied.
\end{itemize}

\subsection{Limitations and Ethics}
\label{sec:limitations}

As hinted in \Cref{sec:model_comparison}, we evaluate models that use supervision from some of the tasks during training.
Specifically, the metric learning models have an advantage on the articulatory distance task.
Nevertheless, the models perform well also on other, more unrelated tasks and we also provide models without this supervision.
We also do not make any distinction between training and development data.
This is for a practical reason because some of the methods we use for comparison are not open embeddings and need to see all concerned words during training.

Another limitation of our work is that we train on phonemic transcriptions, which cannot capture finer grained phonetic distinctions.
Phonemic distinctions may be sufficient for applications such as rhyme detection, but not for tasks such as phone recognition or dialectometry.

We attempted to be inclusive with the language selection and do not foresee any ethical issues.

\bigskip

\section{Future Work}
\label{sec:future_work}

After having established the standardized evaluation suite, we wish to pursue the following: 
\begin{itemize}[noitemsep,topsep=0mm]
\item enlarging the pool of languages,
\item including more tasks in the evaluation suite,
\item contextual phonetic word embeddings,
\item new models for phonetic word embeddings.
\end{itemize}

\nocite{*}
\vspace*{7mm}
\section{Bibliographical References}
\vspace*{-7mm}

\bibliographystyle{misc/lrec_natbib}
\bibliography{misc/bibliography.bib}

\appendix

\end{document}